\documentclass{llncs}  

\usepackage{graphicx} 
\usepackage[T1]{fontenc}
\usepackage{amsmath}
\usepackage[english]{babel}
\usepackage{array,tabularx}
\usepackage{booktabs} 
\usepackage{makeidx}  
\usepackage{xparse}
\usepackage{moredefs}
\usepackage{lips}
\usepackage{epstopdf} 
\usepackage{color}
\usepackage{csquotes}
\usepackage{xcolor}
\usepackage{times}
\usepackage[hidelinks]{hyperref}
\usepackage[nolist,nohyperlinks]{acronym}
\usepackage{comment}
\usepackage{paralist}
\usepackage{cleveref}
\usepackage{wrapfig}
\usepackage{multirow}
\usepackage{microtype}
\usepackage{listings}
\usepackage[super]{nth}
\usepackage{natbib}
\usepackage{siunitx}
\usepackage{todonotes}

\usepackage{float}
\floatstyle{plaintop}
\restylefloat{table}

\usepackage[style=base,labelfont=bf,labelsep=period,tableposition=top,font=small]{caption}
\makeatletter
\renewcommand\@biblabel[1]{#1.}
\makeatother
\addto\captionsenglish{}

\usepackage{fancyhdr}
\fancypagestyle{WI_footer}{\fancyhf{}\fancyfoot[L]{\color{black}\scriptsize 19th International Conference on Wirtschaftsinformatik\\September 2024, Würzburg, Germany}}

\crefformat{footnote}{#2\footnotemark[#1]#3}
\expandafter\def\expandafter\UrlBreaks\expandafter{\UrlBreaks
  \do\a\do\b\do\c\do\d\do\e\do\f\do\g\do\h\do\i\do\j%
  \do\k\do\l\do\m\do\n\do\o\do\p\do\q\do\r\do\s\do\t%
  \do\u\do\v\do\w\do\x\do\y\do\z\do\A\do\B\do\C\do\D%
  \do\E\do\F\do\G\do\H\do\I\do\J\do\K\do\L\do\M\do\N%
  \do\O\do\P\do\Q\do\R\do\S\do\T\do\U\do\V\do\W\do\X%
  \do\Y\do\Z}

\newcolumntype{L}[1]{>{\raggedright\arraybackslash}p{#1}}   
\newcolumntype{C}[1]{>{\centering\arraybackslash}p{#1}}     
\newcolumntype{R}[1]{>{\raggedleft\arraybackslash}p{#1}}    

\begin{document}
\frontmatter          
\pagestyle{headings}  

\mainmatter              

\title{Hate Speech and Sentiment of YouTube Video Comments From Public and Private Sources Covering the Israel-Palestine Conflict}
\subtitle{Research Paper} 

\author{Simon Hofmann\inst{1} \and
Christoph Sommermann\inst{1} \and
Mathias Kraus\inst{2} \and
Patrick Zschech\inst{3} \and
Julian Rosenberger\inst{2}}

\institute{
Friedrich-Alexander Universität Erlangen-Nürnberg, Lange Gasse 20, 90403 Nürnberg\\ 
\email{{simon1995.hofmann, christoph.sommermann}@fau.de} \and
Universität Regensburg, Bajuwarenstraße 4, 93053 Regensburg\\
\email{{mathias.kraus, julian.rosenberger}@ur.de} \and
Universität Leipzig, Grimmaische Straße 12, 04109 Leipzig\\
\email{patrick.zschech@uni-leipzig.de}}

\maketitle
\setcounter{footnote}{0}

\begin{abstract}
This study explores the prevalence of hate speech (HS) and sentiment in YouTube video comments concerning the Israel-Palestine conflict by analyzing content from both public and private news sources. The research involved annotating \num{4983} comments for HS and sentiments (neutral, pro-Israel, and pro-Palestine). Subsequently, machine learning (ML) models were developed, demonstrating robust predictive capabilities with area under the receiver operating characteristic (AUROC) scores ranging from 0.83 to 0.90. These models were applied to the extracted comment sections of YouTube videos from public and private sources, uncovering a higher incidence of HS in public sources (40.4\%) compared to private sources (31.6\%). Sentiment analysis revealed a predominantly neutral stance in both source types, with more pronounced sentiments towards Israel and Palestine observed in public sources. This investigation highlights the dynamic nature of online discourse surrounding the Israel-Palestine conflict and underscores the potential of moderating content in a politically charged environment.\\

\end{abstract}

\thispagestyle{WI_footer}


\section{Introduction}
\label{sec:introduction}

In the digital era, the prevalence of hate speech (HS) on online platforms has notably escalated, particularly within the discourse surrounding contentious socio-political issues. A vivid example is the Israel-Palestine conflict, where the region's complex and longstanding geopolitical tensions frequently manifest in heated online interactions that can escalate into HS \citep{caron2022towards, fortuna2018survey}. The impact of such speech extends beyond mere words, affecting not only the psychological health of individuals but also the overall harmony within their social communities\citep{maarouf_virality_2024, frenda_killing_2023}. In 2021, approximately 40\% of U.S. residents reported encountering online hate speech, with young adults showing heightened vulnerability to its psychological consequences \citep{vogels2021state, saha2019prevalence}. Against this backdrop, our study focuses on HS detection and sentiment analysis within the context of the Israel-Palestine conflict. HS detection refers to the identification and classification of language that denigrates individuals or groups based on inherent characteristics such as race, religion, or gender \citep{loebbecke_ai_2021, fortuna2018survey}. Sentiment analysis assesses the emotional tone behind a text to understand the attitudes, opinions, and emotions being expressed \citep{mukkamala_champions_2023, xie_understanding_2022}. By integrating these analytical dimensions, our research aims to provide deeper insights into the nature of social media discourse, particularly in politically sensitive contexts.

Research underscores that platforms like YouTube not only reflect but can amplify ethno-racial and religious hatred, potentially transforming online spaces into arenas for aggressive and polarized exchanges \citep{Ray_George_2021a, susarla2012social}. This growing prevalence of HS, intensified by the interconnected nature of online platforms, underscores an urgent need for sophisticated machine learning (ML) methods to support more effective content moderation, implement legal regulation, and enhance digital safety \citep{caron2022towards, slivko_regulation_2021}.

This paper addresses the dual challenges of detecting HS and analyzing sentiment in the aftermath of significant events like the Hamas attack on Israel on the 7th of October in 2023 \citep{barnea2024israeli}. By analyzing comments from YouTube videos related to this conflict, we aim to develop ML models that not only detect HS but also gauge public sentiment, thereby contributing to the mitigation of online toxicity and promoting digital responsibility \citep{wlomert2024frontiers}. Our study makes several contributions. First, we provide a carefully annotated dataset of \num{4983} YouTube comments labeled for HS and sentiment.\footnote{Our dataset is publicly available here: \href{https://doi.org/10.17605/OSF.IO/Q45CT}{doi.org/10.17605/OSF.IO/Q45CT}.} Second, we develop ML models with promising predictive capabilities with AUROC scores between 0.83 to 0.90 for tasks of hate speech detection and sentiment analysis. Finally, through this work, we gain novel insights into how different platforms may influence user interactions and the broader dynamics of online discourse \citep{wlomert2024frontiers}.

This paper is structured as follows: Section \ref{sec:related_work} reviews relevant literature to establish a theoretical foundation for our methods. Section \ref{sec:methods} details the methodologies employed in data collection, cleaning, labeling, and analysis. Section \ref{sec:results} presents our findings, followed by a discussion in Section \ref{sec:discussion} on the broader impacts of these results on online discourse moderation. Finally, Section \ref{sec:Limitations} outlines the limitations of our study and suggests directions for future research. This structured approach ensures a comprehensive exploration of the dynamics of HS and sentiment within the context of the Israel-Palestine conflict, offering clear insights and actionable findings.

\section{Related Work}
\label{sec:related_work}

The increasing prevalence of HS on online platforms, particularly during discussions on sensitive topics like the Israel-Palestine conflict, has highlighted the need for effective moderation tools \citep{vosoughi2018spread}. ML techniques have been widely adopted to tackle this challenge, with a focus on developing accurate and efficient HS detection models \citep{caron2022towards}.

Traditional ML approaches, such as Logistic Regression (LR) and Support Vector Machines (SVM), have been extensively used for HS detection and sentiment analysis tasks. LR, a probabilistic model that uses a logistic function to model the relationship between input features and binary outcomes, has shown promising results in text classification tasks \citep{Blümel_Zaki_2022}. SVM, on the other hand, aims to find the optimal hyperplane that maximizes the margin between different classes in a high-dimensional space \citep{cortes1995support}. Both LR and SVM have been successfully applied to HS detection and sentiment analysis in various contexts \citep{fortuna2018survey, schmidt2017survey}.

While more advanced techniques like deep learning and Transformer architectures have gained popularity in recent years \citep{siebers_survey_2022}, traditional ML methods still hold value due to their simplicity, interpretability, and effectiveness, especially when dealing with smaller datasets. For instance, \citet{davidson2017automated} employed LR and SVM for detecting toxic language in social media comments, achieving promising results. Similarly, \citet{burnap2015cyber} used LR and SVM to classify hateful and antagonistic content on Twitter, demonstrating the effectiveness of these methods in capturing the nuances of online HS.

Despite these advancements, significant challenges remain in HS detection and sentiment analysis, particularly in managing the complex socio-political discourse surrounding the Israel-Palestine conflict. The nuanced language and evolving nature of online interactions further complicate the automated detection of HS. Moreover, research highlights distinct behaviors among authors of HS, who tend to exhibit higher activity levels, use specific and non-trivial language, and maintain dense connections within their networks \citep{mathew2020hate, ribeiro2018characterizing}. This trend is especially pronounced in the Global South, where social media plays a critical role in spreading fear and animosity towards minorities, often using historical events and political symbols to stir up emotions \citep{chandra2021virus, jafri2023uncovering}.

Future research should focus on refining ML techniques, improving their adaptability to different contexts, and reducing biases that may affect the fairness of HS detection systems \citep{vosoughi2018spread}. In this study, we aim to contribute to the existing body of work by developing LR and SVM models for HS detection and sentiment analysis in the context of the Israel-Palestine conflict. By focusing on these traditional ML approaches and leveraging a carefully annotated dataset of YouTube comments, we seek to provide insights into the dynamics of online discourse and the effectiveness of these methods in capturing the nuances of HS and sentiment in this context.

\section{Methods}
\label{sec:methods}
\begin{figure}[h!]
    \centering    \includegraphics[width=1\textwidth]{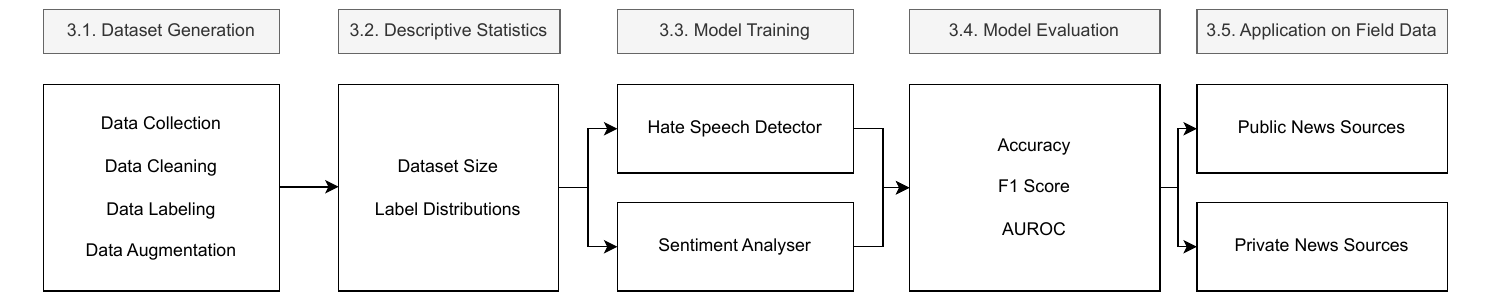}
    \caption{Framework of our work.}
    \label{fig:framework}
\end{figure}

In the following, we describe the steps taken in order to build a dataset, train models, and then apply these models on field data. Figure \ref{fig:framework} illustrates these steps.

\subsection{Dataset Generation}
\label{sec:data_preparation}

We created a unique dataset for this research by collecting comments from YouTube videos related to the Israel-Palestine conflict using web scraping techniques \citep{mitchell2018web}. We gathered comments from videos of German public and private news services to ensure diverse perspectives. The data collection process was conducted in two rounds, spanning seven weeks from October 2 to November 20, 2023. In the first round, we collected a total of \num{19047} comments from 13 videos. To capture the evolving nature of the conflict and its impact on public discourse, we carried out a second scraping process to obtain comments from an additional six videos that covered events occurring on a weekly basis after October 18, 2023.

Given the noisy nature of user-generated content on YouTube, we extensively cleaned the collected dataset to ensure data quality and consistency. We removed empty lines, duplicates, non-German comments, numbers, special characters, and emojis. We standardized the text by converting it to lowercase, normalizing umlauts and special letters, and applying tokenization. We utilized the Natural Language Toolkit (NLTK) to filter out stop words while retaining negations for sentiment analysis \citep{bird2006nltk}. We employed lemmatization using the WordNet Lemmatizer from NLTK to reduce words to their base form, accounting for the complexities of the German language \citep{plisson2004rule}.

To train supervised machine learning models for hate speech detection and sentiment analysis, we manually labeled the cleaned dataset. Three annotators conducted a detailed and iterative labeling process, following recommendations by \citet{artstein2008inter} to ensure consistency and reliability. We used a binary system to label comments for the presence of hate speech, with an expanded definition including derogatory, sexist, vulgar, or profane language. For sentiment labeling, we employed a ternary classification (pro-Palestine, pro-Israel, and neutral) based on keywords, phrases, and overall tone. Regular meetings among annotators ensured consistent application of labeling criteria.

To mitigate class imbalance and increase dataset diversity, we employed data augmentation techniques. We adapted the "HateCheck" English dataset \citep{rottger2020hatecheck} into German through back-translation, enhancing linguistic variety. We also utilized language models, particularly GPT-4, to generate additional comments for sentiment analysis. The augmented dataset included \num{2000} comments for hate speech detection and \num{1095} comments for sentiment analysis, increasing robustness for model training \citep{yang2023getting}. Table \ref{tab:data_augmentation} summarizes the data augmentation techniques and their contributions.

\begin{table}[htbp]
\centering
\caption{Summary of data augmentation techniques.}
\label{tab:data_augmentation}
\begin{tabular}{@{}llr@{}}
\toprule
Data augmentation technique & Source & Comments added \\
\midrule
Back Translation & HateCheck via DeepL & \num{2000} \\
GPT-4 & Generated Content & \num{1095} \\
\bottomrule
\end{tabular}
\end{table}

\subsection{Descriptive Statistics}
\label{sec:descriptive_stats}

Following the preprocessing, two distinct datasets emerged. The initial dataset, comprised of the original scraped comments, included \num{19074} entries. From this dataset and the augmented comments, a total of \num{4983} comments were manually classified for HS and sentiment by the annotators.
The subsequent, more concise dataset evolved from the nearly five thousand annotated comments, encapsulating 944 distinctly labeled entries. The distribution between public (55.08\%, n=520) and private (44.92\%, n=424) sources was noted, with the sampled videos spanning seven weeks from October 2 to November 20, 2023. A majority of the comments (62.12\%, n=577) originated from the initial three weeks up to October 18, 2023, while the remainder (38.88\%, n=367) came from the concluding four weeks. Comment length varied widely within the dataset, enriching it for sentiment analysis and subsequent model training.

In labeling, efforts were concentrated on achieving label balance. The label distribution shows a higher prevalence of HS (64.55\% Yes vs. 35.45\% No), which may slightly bias the model towards detecting HS, potentially increasing false positives. Sentiment distribution is predominantly neutral (46.39\%), with more pro-Palestine (33.15\%) than pro-Israel (20.45\%) comments, which may affect the model’s sentiment detection accuracy. The relatively balanced public/private source distribution and temporal spread of comments help ensure diverse perspectives, though the higher initial activity may influence overall representativeness. The distribution of labels for both HS and sentiment is detailed in Table \ref{tab:label_distribution}.

\begin{table}[htbp]
\centering
\caption{Distribution of hate speech and sentiment labels in the annotated dataset.}
\label{tab:label_distribution}
\begin{tabular}{lC{1cm}C{1cm}p{1cm}lC{1cm}c}
\toprule
Label & Count & Percentage & & Label & Count & Percentage \\
\midrule
\textbf{Hate Speech} & & & & \textbf{Sentiment} & & \\[0.3em]
Yes & \num{609} & \num{64.55} & &  Neutral & \num{438} & \num{46.39} \\
No & \num{335} & \num{35.45} & & Pro-Israel & \num{193} & \num{20.45} \\
& & & & Pro-Palestine & \num{313} & \num{33.15} \\
\bottomrule
\end{tabular}
\end{table}

\subsection{Model Training}
\label{sec:machine_learning_models}

Machine learning has been widely used for hate speech detection and sentiment analysis in recent years, with various techniques employed to represent text data and train classification models \citep{fortuna2018survey, Thoms_Eryilmaz_Mercado_Ramirez_Rodriguez_2017}. Our decision to combine hate speech detection with sentiment analysis is grounded in several key considerations from existing literature. \citet{schmidt2017survey} highlight that sentiment analysis can provide valuable context for hate speech detection, as extreme negative sentiment often correlates with hate speech. This complementary relationship allows for a more nuanced understanding of the text. Moreover, \citet{davidson2017automated} argue that sentiment analysis helps in distinguishing between hate speech and merely offensive language, which often share negative sentiment but differ in intent and severity. By integrating these two tasks, we aim to provide a more comprehensive analysis of the YouTube comments related to the Israel-Palestine conflict, leveraging the strengths of both approaches to gain deeper insights into the nature and tone of the discourse.


We experimented with two methods for text representation: Bag of Words (BoW) and Term Frequency-Inverse Document Frequency (TF-IDF) \citep{siebers_survey_2022}. BoW represents text as a vector of word frequencies, disregarding grammar and word order but keeping multiplicity. This method is simple and effective for converting text into a numerical format for machine learning models \citep{zhang2010understanding}. TF-IDF, on the other hand, weights the frequency of terms inversely with their occurrence across documents, highlighting unique words within the corpus \citep{ramos2003using}. We opted for BoW due to its simplicity and effective performance during our initial experimentation phase.

For the hate speech detection task, we trained separate LR and SVM models. The LR model used the 'lbfgs' solver, which is an optimizer in the family of quasi-Newton methods. The regularization strength was set to 0.1 to mitigate potential overfitting issues, and the L2 norm was employed for regularization. This configuration provided a good balance between bias and variance, ensuring the model's robustness and reliability. The SVM model for hate speech detection used a linear kernel function, as it is well-suited for this classification problem \citep{cortes1995support}. The hyperparameter tuning for the SVM involved testing a range of values for the regularization parameter from 0.01 to 1.0 to determine the best performing configuration.

Similar to the hate speech detection task, we trained separate LR and SVM models for sentiment analysis. The LR model for sentiment analysis also used the 'lbfgs' solver and L2 regularization, with the regularization strength set to 0.1. The SVM model for sentiment analysis employed a linear kernel function, and the hyperparameter tuning process was conducted in the same manner as for the hate speech detection SVM model.

By training separate models for hate speech detection and sentiment analysis, we developed a Hate Speech Detector and a Sentiment Analyser. This approach allows for a more targeted and specialized analysis of the YouTube comments, enabling us to gain insights into the prevalence of hate speech and the sentiment expressed towards the Israel-Palestine conflict.

\subsection{Model Evaluation}
\label{sec:evaluation}
To  evaluate our ML models, we followed common practice. The prepared datasets (see Sections \ref{sec:data_preparation}-\ref{sec:descriptive_stats}) were divided into training (80\%) and testing (20\%) sets to facilitate the evaluation of the models' performance on unseen data. 10-fold cross-validation was implemented on the training data to provide a comprehensive view of the models' predictive capabilities and to identify any biases or vulnerabilities in their performance \citep{kohavi1995study}. 

For measuring the model performance, we used various threshold dependent and threshold independent metrics. First, we used the prediction accuracy to obtain first insights into the performance of the ML models. Second, we used F1-score to address the issue of our unbalanced dataset. Finally, we used the are under the receiver operating characteristic (AUROC) as a threshold independent metric.

\subsection{Application on Field Data}
\label{sec:field_analysis}

To assess the practical applicability of our developed hate speech detection and sentiment analysis models, we applied our trained models on two large datasets comprising YouTube comments. The first dataset, consisting of \num{19074} comments, partially served as training material for the models, while the second dataset, containing \num{12000} entirely new comments, was used to apply the model to unfamiliar data in order to gain insights into the political discourse and the dynamics behind it. For each comment, our trained models are used to detect hate speech and evaluate the sentiment expressed.

\begin{figure}[htp]
    \centering    
    \includegraphics[width=0.5\textwidth]{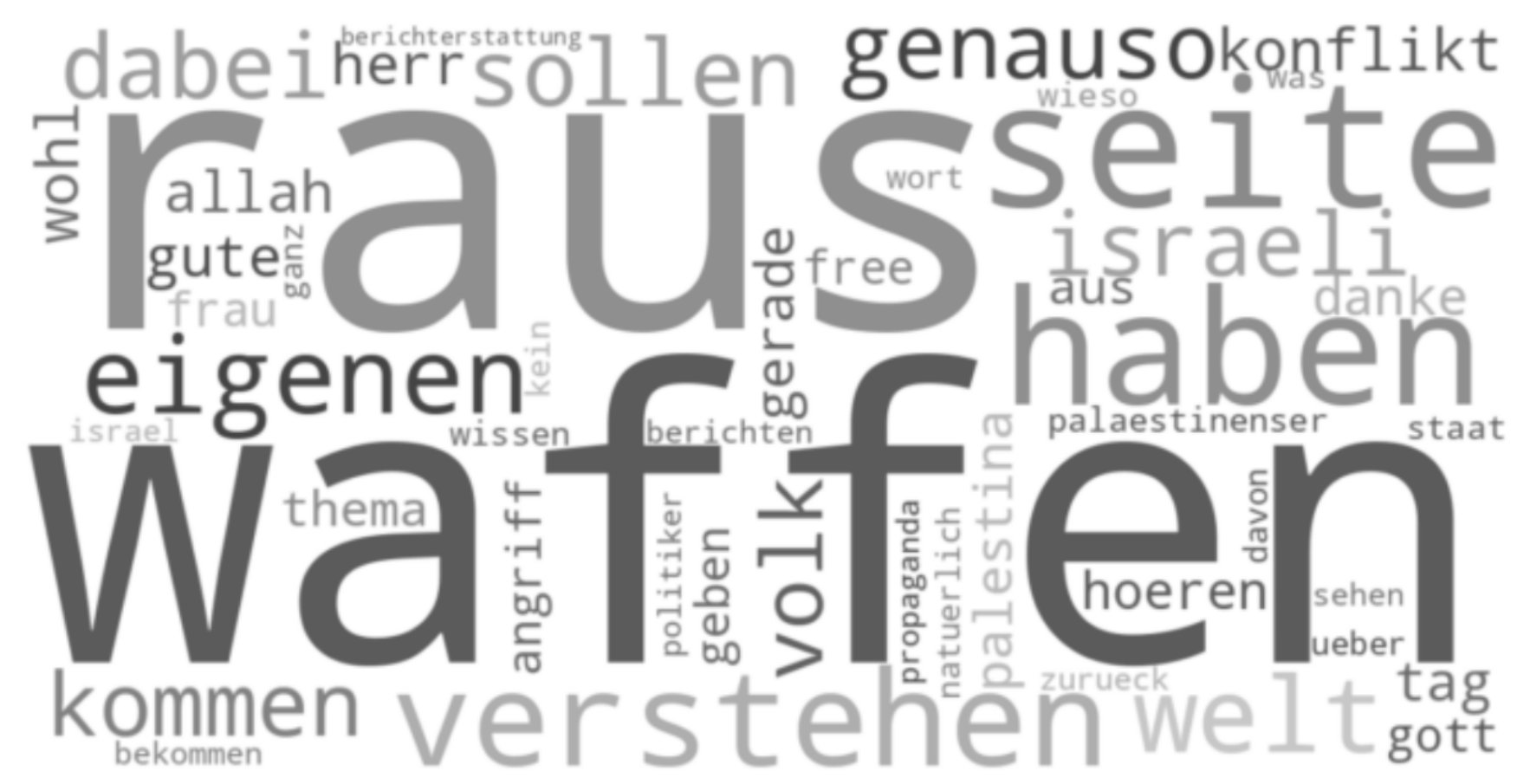}
    \caption{Word cloud of field data.}
    \label{fig:word_cloud}
\end{figure}

Figure \ref{fig:word_cloud} shows a word cloud to visualize the most common words that appear in the field data, where the size of the words represents their occurrence, offering a visual representation of key terms from our data. In this analysis, certain words such as 'raus' and 'waffen' stand out due to their size, each appearing in 1.5 percent of the comments. The most dominant term, 'raus,' a German word meaning 'out' or 'outside', suggests themes of expulsion, removal, or a call to action to leave, which are central in the text. This is confirmed by other prominently featured words like 'volk', 'waffen', 'eigenen', and 'haben', indicating discussions around national identity, defense, and possession. Additionally, words such as 'israeli', 'allah', and 'volk' point to the Israeli–Palestine conflict. Smaller, yet still significant terms like 'verstehen', 'thema', and 'berichterstattung' imply a focus on understanding reports or discussions. It is important to remember that word clouds, while providing a snapshot of frequency, do not offer context or the relationships between terms. They guide us toward the main subjects and the possible tone or stance but do not replace thorough textual analysis. 

\section{Results}
\label{sec:results}

\subsection{Hate Speech Detection}
\label{sec:results_hate_speech}

The evaluation results of our HS detection model using LR present a comprehensive picture of its performance. It yielded a training accuracy of 77.46\% and a testing accuracy of around 73.94\%. While lower than the testing accuracy, this figure is still indicative of good model reliability. The classification report illustrates precision, recall, and F1-scores of about 0.74 for both classes. Specifically, for class 0 (No HS), the precision, recall, and F1-score were 0.74, 0.71, and 0.72, respectively, while for class 1 (HS), these metrics were 0.74, 0.76, and 0.75, respectively. The macro and weighted average F1-scores were both approximately 0.74, indicating a balanced performance across classes. A key metric in classification tasks, AUROC, stood at 0.83 for both classes. These results underscore the effectiveness of the LR model in the context of our HS detection task. The careful tuning of model parameters, such as the regularization strength, has contributed to this solid performance, helping to mitigate overfitting while maintaining the model's ability to generalize well to new, unseen data. 

The SVM model achieved an AUROC score of 0.82 for both classes, slightly performing worse than the LR. The confusion matrix shows that the model correctly classified 225 instances as negative and 263 instances as positive, while misclassifying 91 negative instances as positive and 81 positive instances as negative. The results for both models are close together. Based on the slightly better values of the AUROC and in the confusion matrix it was decided to continue with the LR model and apply it to the field data for hate speech detection.

\subsection{Sentiment Analysis}
\label{sec:results_sentiment_analysis}

The LR model for the sentiment analysis achieved an accuracy of 76.92\% on the training set, while on the testing set, it maintained a relatively strong performance with an accuracy of 72.06\%. Furthermore, the model's discrimination ability across classes was highlighted by the AUROC scores. For class 0 (Neutral), the AUROC was 0.90, indicating robust predictive power. Class 1 (Israel) exhibited an AUROC of 0.85, suggesting reliable discrimination, while class 2 (Palestine) showed a respectable AUROC of 0.84. These findings underscore the model's capability to differentiate between classes relatively well.

\subsection{Field Analysis}
\label
{sec:results_field_analysis}

\subsubsection{Hate Speech Detection.}

The comparison between the incidence of HS in private and public domains reveals that in private sources, a smaller proportion of content is classified as hate (31.6\%), while a larger portion (68.4\%) is categorized as no hate. In public sources, the proportion of content identified as hate is slightly higher at 40.4\%, with 60.2\% being no hate. These numbers suggest that HS is more frequently detected in public sources than in private ones according to this analysis. However, it is important to consider the context of the data, the methodology used for detection, and the definitions applied to what constitutes HS. 

\begin{figure}[h!]
    \centering    
    \includegraphics[width=0.8\textwidth, trim={0.1cm 0.1cm 0.1cm 0.1cm}, clip]{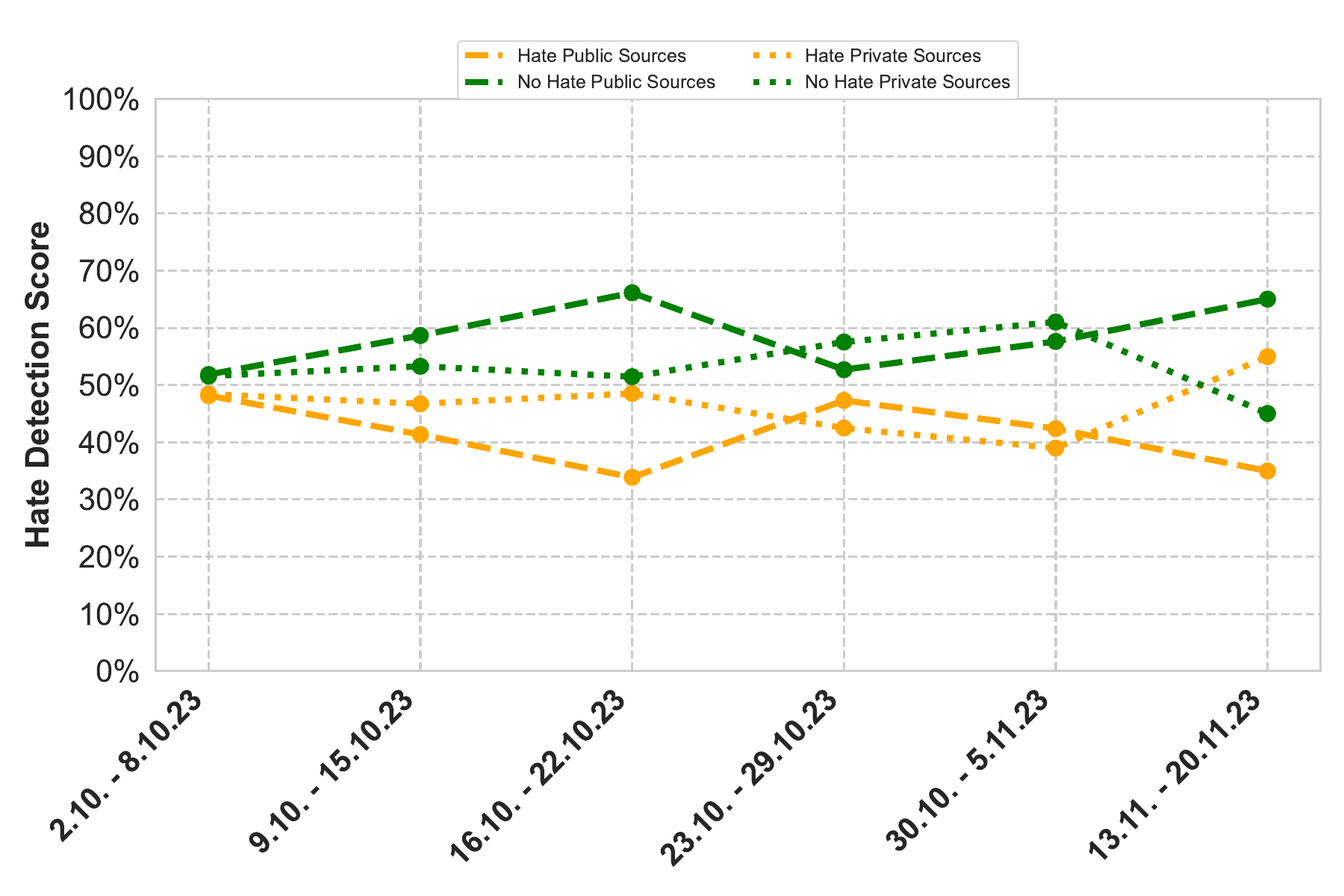}
    \caption{Hate progression over time.}
    \label{fig:hate_progression}
\end{figure}

Figure \ref{fig:hate_progression} shows four lines tracking the percentage of content identified as hate versus no hate over time. The hate lines fluctuate over time, indicating variability in the comments of public and private sources during the specified periods. There appears to be a general difference of no hate content being higher than hate content across all the dates shown. The dates on the x-axis range from October 2 to November 20, 2023.  The highest peak for the hate line, representing public sources, occurs in the week of '02.10. - 08.10.23', which may be related to the onset of escalated military actions and hostilities, potentially contributing to a surge in online hate speech as communities reacted to the sudden escalation and heavy casualties \citep{NewYorkTimes2023a}. Accordingly, the lowest point for the hate line from public sources is seen in the week of '16.10. - 22.10.23', which could align with periods of relative calm or effective diplomatic interventions that temporarily moderated discussions \citep{CNN2023a}. The no hate lines also fluctuate, but in an inverse relationship to the hate line. The reasons for these fluctuations could be influenced by many factors such as current events or changes in public discourse. These fluctuations highlight the dynamic nature of online content moderation and the challenges involved in monitoring HS over time.

Figure \ref{fig:hate_progression} further shows fluctuations in the detection of HS in private sources over time. The green lines representing no HS in private and public sources shows a wave-like pattern with peaks and troughs. The highest peak for HS occurs in the week of '13.11. - 20.11.23', where the detection rate in private sources rises to just under 55\%. This peak likely reflects the public reaction to ongoing military operations and particularly impactful events, such as airstrikes on hospitals and densely populated areas reported during this week, which could drive spikes in hateful and polarized online content \citep{BBC2023a, NewYorkTimes2023b}. The hate detection rate for private sources reaches its lowest point in the fifth week, '30.10. - 05.11.23', at around the 40\% mark. This dip may be associated with brief periods of ceasefire or calls for peace that could lead to a temporary reduction in online hostility \citep{CNN2023b}. The green line, representing no HS, and the orange line, representing HS, both of which representing public and private sources, intersect before the week of '13.11. - 20.11.23'. At this point, the rates of hate and no hate detection are equal, indicating a balance between detected HS and non-HS content. As with public sources, the reasons behind these fluctuations in private sources are not provided by the graph alone. These data points underscore the complexity of tracking and understanding HS patterns within different domains of communication.

\subsubsection{Sentiment Analysis.}

A majority of the content from private sources is categorized as neutral, comprising 68.4\%. In private sources, sentiment towards Israel accounts for 18.6\%, while sentiment towards Palestine is at 13\%. Conversely, a slightly lower percentage of content from public sources is neutral, at 63.6\%. Public sources exhibit a higher sentiment towards Israel at 21.2\%, and sentiment towards Palestine is also more pronounced at 15.2\%. Neutral sentiment still dominates in both private and public sources, suggesting that most content does not express a strong bias towards either Israel or Palestine. Notably, sentiments towards Israel and Palestine are more elevated in public sources than in private ones. The lesser prevalence of neutral sentiment in public sources could suggest that public platforms or media may feature more opinionated or specific content related to Israel and Palestine. 

\begin{figure}[h!]
    \centering    
    \includegraphics[width=0.8\textwidth, trim={0.1cm 0.1cm 0.1cm 0.1cm}, clip]{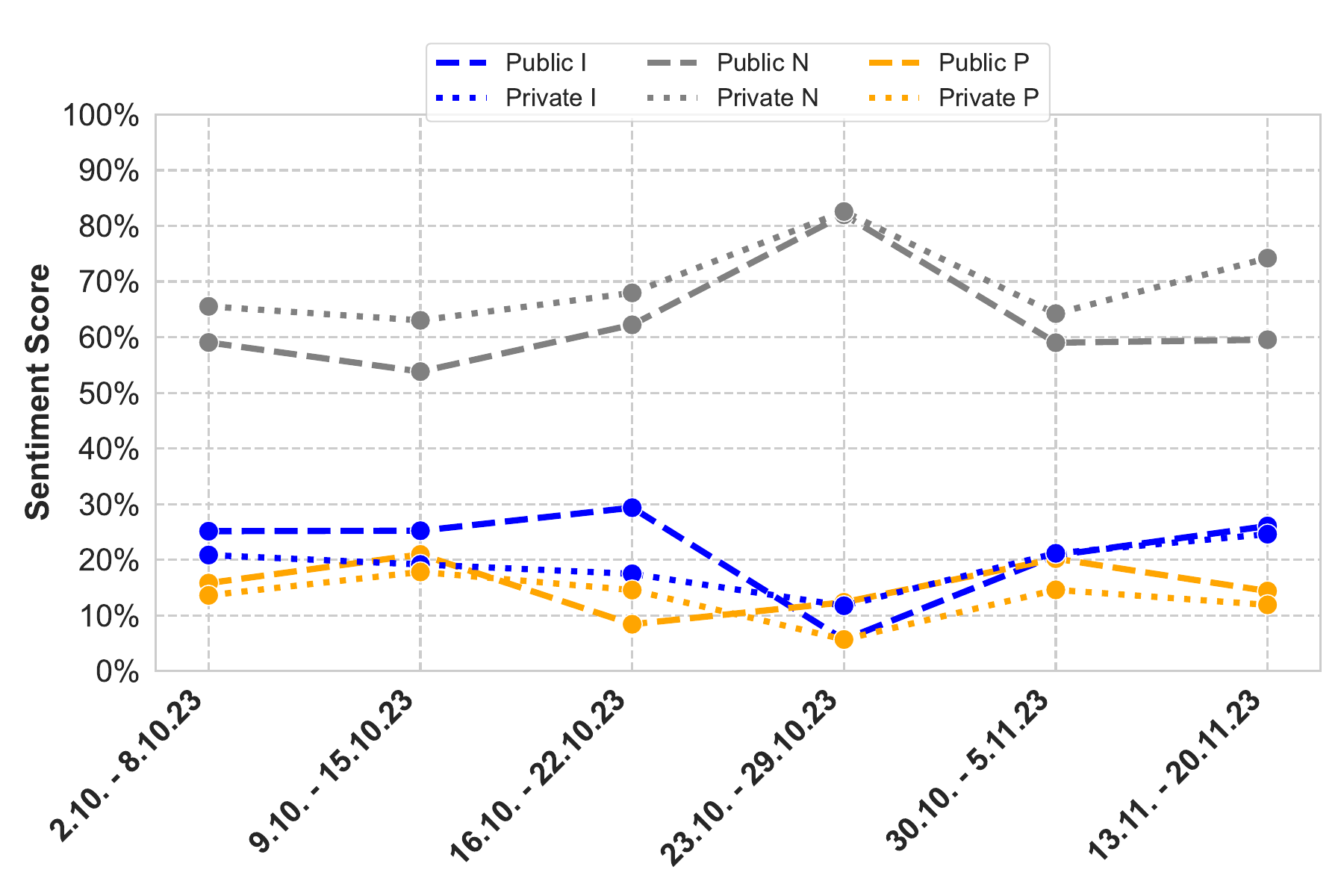}
    \caption{Sentiment progression over time.}
    \label{fig:sentiment_progression}
\end{figure}

Figure \ref{fig:sentiment_progression} illustrates the distribution of sentiment across private and public sources on the topic of Israel and Palestine, categorized into three key sentiments: Neutral (N), representing comments that neither support nor oppose either side; pro-Israeli (I), representing comments that support or favor Israel; and pro-Palestinian (P), representing comments that support or favor Palestine. The graph shows six lines representing these sentiment categories across sources: Neutral sentiment is shown in gray; pro-Israeli sentiment is shown in blue; pro-Palestinian sentiment is shown in orange. The difference between public and private news sources is represented by the dotted line for private sources and the dashed line for public sources. This visualization helps to understand how each sentiment fluctuates over time and varies between public and private contexts. 

Neutral sentiment generally peaks at around 80\% by the end of October across both private and public sources, though it shows slight fluctuations before and afterwards. Sentiment towards Israel in public sources starts just above 20\% in early October, experiences a dip towards the end of the month, and then rises again, occasionally surpassing initial levels. Conversely, sentiment towards Palestine in public sources begins below 20\%, increases in early October, then dips below its initial level in the third week to around 10\%, before climbing towards the end of the month. The orange dotted line, representing pro-Palestinian sentiment in private sources, shows a fluctuating pattern, starting at 13.61\% in early October, peaking at 17.82\% in mid-October, before a significant drop to 5.68\% at the end of the month, and finally diminishing to a mere 1.19\% by mid-November. This suggests a marked decrease in private pro-Palestinian sentiment as the month progresses. In contrast, the orange dashed line, which charts pro-Israeli sentiment in public sources, begins at 15.81\%, peaks notably to 20.96\% in mid-October, then dips to 8.42\% before recovering back to 20.20\% at the start of November, and stabilizes at 14.38\% by mid-November, indicating a more resilient public sentiment towards Palestine throughout the period. Despite these fluctuations, neutral sentiment remains consistently the highest, indicating that the majority of content does not express a strong bias towards either Israel or Palestine. This graph demonstrates similar patterns of sentiment movement in both private and public sources, with fluctuations particularly pronounced in private sources, where neutral sentiment notably spikes around October 23rd, and sentiments towards Israel and Palestine demonstrate similar trends of increase and decrease, concluding the month with noticeable shifts.

\section{Discussion}
\label{sec:discussion}

The analysis of online discourse from YouTube comments over a seven-week period concerning the Israel-Palestine conflict, which encompassed \num{19074} comments with \num{4983} specifically annotated for HS and sentiment, provides an insightful look into the complex dynamics of public sentiment on digital platforms. This period, marked by intensified geopolitical tensions, saw an interplay of opinions that the application of sentiment analysis and HS detection has helped to decode, echoing findings from previous research that highlights how negative sentiments typically dominate direct content interactions \citep{oh2023you}. Furthermore, Figure \ref{fig:word_cloud}, featuring terms like 'palestina', 'israeli', 'free', and 'god', highlights the conflict’s historical, territorial, and nationalistic core, as well as its deep religious dimensions. The term 'free' especially highlights the ongoing struggles for independence, sovereignty, and liberation, which are central to the geopolitical discourse on the Israel-Palestine conflict.
Studies like \citet{oh2023you} highlight that while negative sentiment often pervades direct interactions, positive sentiment tends to spread more broadly across social platforms. This understanding of sentiment distribution is critical because it informs the strategies that can be employed to effectively moderate and engage with content. 

The predominance of neutral sentiments in both private and public sources, as revealed in this study, suggests a general reluctance among users to openly declare their stance on such a controversial issue. However, the analysis also reveals that pro-Palestine sentiments, even within comments classified as HS, are notably vocal, sometimes crossing into HS territory. This observation is crucial for content moderators as it highlights the importance of nuanced and balanced labeling practices to avoid mistakenly conflating strong sentiments with HS. It is imperative that short comments like "free Palestine" are carefully scrutinized to prevent oversimplification, while longer comments, which offer more context, require advanced analytical tools to accurately discern underlying sentiments. In addition, the data suggests significant fluctuations in sentiment that may be influenced by specific external events, be they political, military, or social. These fluctuations indicate a potential impact of real-world events on online expression and highlight the possibility for sentiment analysis to serve as a barometer for public opinion and discourse dynamics. Such insights could be crucial not only for developing more effective communication strategies but also for potentially enhancing our understanding of how global events may shape online communities. 

This comprehensive analysis highlights the critical need for robust HS detection mechanisms, particularly in discussions on contentious topics like the Israel-Palestine conflict. By refining these tools and approaches, we can cultivate constructive online environments and maintain digital platforms for balanced, informed discussions, not fostering hostility. Implementing moderation strategies, such as warning labels and sentiment analysis alerts, can balance debate while flagging harmful content for review.

\subsection{Limitations and Further Research}
\label{sec:Limitations}

This study has several limitations. Firstly, scraping comments from 13 videos, mainly from public and private channels, introduces uniqueness that may affect reproducibility when using different selection criteria. Additionally, disabled comments on some videos limited data access, potentially impacting model outcomes. Also, training the models on a limited dataset could affect their robustness and applicability. Furthermore, during data cleaning, emojis were removed due to their frequent sarcastic use, which complicates sentiment and hate speech interpretation. This may limit analysis by excluding non-textual expressions that provide additional context. This study focused on descriptive analysis to provide a clear overview of the data, while a thorough exploration of underlying reasons requires a separate study. Moreover, analyzing user demographics was not feasible with the available data, as the YouTube Data API does not provide this information. The research setting, including the time, location, and researchers' preferences, adds to the peculiarities and uniqueness of this study, thus potentially limiting the generalizability of the findings. Future research could adopt a longitudinal approach to studying the Israel-Palestine conflict, analyzing sentiment over an extended period to capture evolving public opinion more accurately. Increasing the proportion of precisely labeled comments and further refining the models could provide deeper insights into which sources or topics are most associated with HS, aiding the development of specialized HS detection models. Such efforts could help make online interactions more pleasant by effectively mitigating HS and abusive language on various platforms.
Additionally, the researchers' German background influenced the choice of German videos for analysis, ensuring highest linguistic proficiency. This study focuses solely on analyzing online discourse and does not address potential biases from a dominant German perspective on the conflict, which is a limitation. Lastly, this study does not analyze the content or sentiment of the videos themselves, excluding potential insights into how media framing influences the comments.


\bibliographystyle{agsm}
\bibliography{_literature}

\end{document}